\documentclass{article}

\usepackage{arxiv}

\usepackage[utf8]{inputenc} % allow utf-8 input
\usepackage[T1]{fontenc}    % use 8-bit T1 fonts
\usepackage{hyperref}       % hyperlinks
\usepackage{url}            % simple URL typesetting
\usepackage{booktabs}       % professional-quality tables
\usepackage{amsfonts}       % blackboard math symbols
\usepackage{nicefrac}       % compact symbols for 1/2, etc.
\usepackage{microtype}      % microtypography
\usepackage{lipsum}
\usepackage{multirow}
\usepackage{amsmath,amssymb,amsfonts}
\usepackage{xcolor}
\usepackage{algorithm}
\usepackage{algorithmic}
\usepackage{subcaption}
\usepackage{enumitem}
\usepackage{graphicx}
\graphicspath{ {./images/} }

\title{motion-to-Response Content Generation via Multi-Agent AI System with Real-Time Safety Verification}

\author{
 HyeYoung Lee \\
  Department of Artificial Intelligence\\
  Korean University\\
  Seoul, South Korea \\
  \texttt{uohesha@korea.ac.kr}
}

\begin{document}
\maketitle
\begin{abstract}
This paper proposes a multi-agent artificial intelligence system that generates response-oriented media content in real time based on audio-derived emotional signals. Unlike conventional speech emotion recognition studies that focus primarily on classification accuracy, our approach emphasizes the transformation of inferred emotional states into safe, age-appropriate, and controllable response content through a structured pipeline of specialized AI agents. The proposed system comprises four cooperative agents: (1) an Emotion Recognition Agent with CNN-based acoustic feature extraction, (2) a Response Policy Decision Agent for mapping emotions to response modes, (3) a Content Parameter Generation Agent for producing media control parameters, and (4) a Safety Verification Agent enforcing age-appropriateness and stimulation constraints. We introduce an explicit safety verification loop that filters generated content before output, ensuring compliance with predefined rules. Experimental results on public datasets demonstrate that the system achieves 73.2\% emotion recognition accuracy, 89.4\% response mode consistency, and 100\% safety compliance while maintaining sub-100ms inference latency suitable for on-device deployment. The modular architecture enables interpretability and extensibility, making it applicable to child-adjacent media, therapeutic applications, and emotionally responsive smart devices.
\end{abstract}

\textbf{Keywords:} Speech Emotion Recognition, Multi-Agent Systems, Content Generation, Safety Verification, On-Device AI

\section{Introduction}
Audio-based emotion recognition has emerged as a fundamental component in human-centered AI systems, enabling machines to perceive and respond to human affective states. Prior research has demonstrated that emotional cues embedded in speech signals can be effectively captured using deep learning models \cite{ser_survey}, with applications spanning healthcare monitoring, educational technologies, and interactive media systems. However, most existing work terminates at the emotion classification stage, leaving open the critical question of how recognized emotions should be operationalized into meaningful, safe, and contextually appropriate responses.

This limitation becomes particularly pronounced in sensitive application domains such as child-oriented media, elderly care systems, and therapeutic interventions. In these contexts, response generation must satisfy multiple constraints simultaneously: emotional congruence with the detected state, age-appropriateness of content, avoidance of overstimulation, and real-time execution on resource-constrained devices. A system that accurately recognizes sadness but responds with inappropriate or overwhelming content fails to serve its intended purpose.

To address these challenges, we introduce a comprehensive emotion-to-response content generation framework that explicitly separates perception, decision-making, generation, and safety verification into cooperating AI agents. As illustrated in Figure \ref{fig:system_overview}, our system processes audio input through a sequential pipeline where each agent performs a specialized function while maintaining explicit interfaces for interpretability and debugging.

The primary contributions of this work are as follows:

\begin{itemize}[leftmargin=*,nosep]
\item A multi-agent architecture that bridges emotion recognition and response content generation with explicit policy and safety layers.
\item A safety verification mechanism with rule-based constraints ensuring age-appropriateness and controlled stimulation levels.
\item A lightweight implementation suitable for on-device deployment with sub-100ms inference latency.
\item Comprehensive experimental validation demonstrating stable response generation without reliance on sensitive user data.
\end{itemize}
\section{Related Work}

\subsection{Speech Emotion Recognition}

Speech emotion recognition (SER) has been extensively studied using spectral features such as Mel-Frequency Cepstral Coefficients (MFCCs), spectrograms, and prosodic features \cite{mfcc_emotion}. Deep neural architectures including Convolutional Neural Networks (CNNs) \cite{cnn_ser}, Recurrent Neural Networks (RNNs) \cite{lstm_emotion}, and attention-based models \cite{attention_ser} have achieved significant improvements on benchmark datasets.

Public datasets such as IEMOCAP \cite{iemocap}, RAVDESS \cite{ravdess}, and EMO-DB \cite{emodb} have enabled standardized evaluation across emotion categories. Recent work has explored cross-corpus generalization \cite{cross_corpus} and multimodal fusion \cite{multimodal_ser}. Despite these advances, SER research typically evaluates success using classification accuracy alone, without considering how recognized emotions translate into downstream applications.

\subsection{Affective Content Generation}

Affective computing research has explored emotion-aware content adaptation, often leveraging generative models for music \cite{music_generation}, speech synthesis \cite{emotional_tts}, and visual content \cite{affective_image}. Large language models have been applied to emotionally responsive dialogue systems \cite{empathetic_dialogue}.

However, these approaches frequently lack explicit safety constraints and may generate content that is contextually inappropriate or overly stimulating. Furthermore, many systems rely on large-scale user interaction data, limiting their applicability in privacy-sensitive domains such as child-oriented applications.

\subsection{Multi-Agent AI Systems}

Multi-agent systems decompose complex tasks into modular components, enabling specialization and explicit reasoning \cite{multiagent_book}. Recent applications include cooperative game playing \cite{marl_games}, robotic coordination \cite{multiagent_robots}, and conversational AI \cite{agent_dialogue}.

In the context of emotion-aware systems, multi-agent architectures remain underexplored. Existing work typically employs monolithic models that jointly perform recognition and generation, sacrificing interpretability and controllability. Our work addresses this gap by introducing a structured multi-agent pipeline with explicit safety verification.
\section{System Architecture}

\subsection{Overview}
\begin{figure*}[t]
\centering
\includegraphics[width=0.95\textwidth]{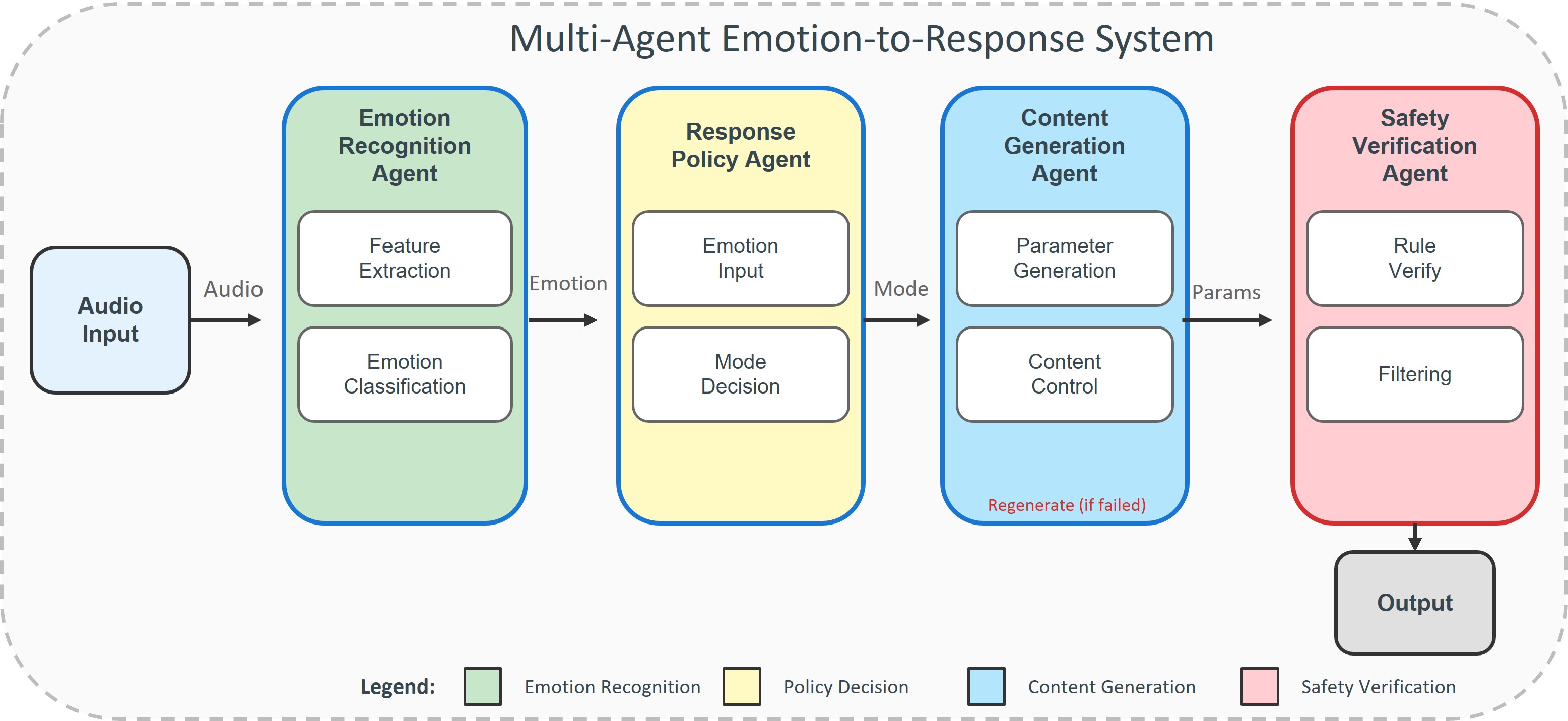}
\caption{Overall architecture of the proposed multi-agent system for emotion-to-response content generation. The system comprises four specialized agents operating sequentially: Emotion Recognition Agent, Response Policy Decision Agent, Content Parameter Generation Agent, and Safety Verification Agent. Failed safety verification triggers content regeneration.}
\label{fig:system_overview}
\end{figure*}
The proposed system comprises five functional components operating in a sequential pipeline, as shown in Figure \ref{fig:system_overview}:

\begin{enumerate}[leftmargin=*,nosep]
\item \textbf{Audio Input Module}: Receives audio signals from microphones or audio sensors.
\item \textbf{Emotion Recognition Agent}: Extracts acoustic features and estimates emotional state.
\item \textbf{Response Policy Decision Agent}: Maps emotional state to a discrete response mode.
\item \textbf{Content Parameter Generation Agent}: Produces numerical parameters controlling media characteristics.
\item \textbf{Safety Verification Agent}: Enforces rule-based constraints before output.
\item \textbf{Content Output Module}: Delivers verified content to rendering systems.
\end{enumerate}

This modular design enables independent optimization of each component, explicit interfaces for debugging, and flexible deployment across different hardware configurations.

\begin{figure}[t]
\centering
\includegraphics[width=0.75\columnwidth]{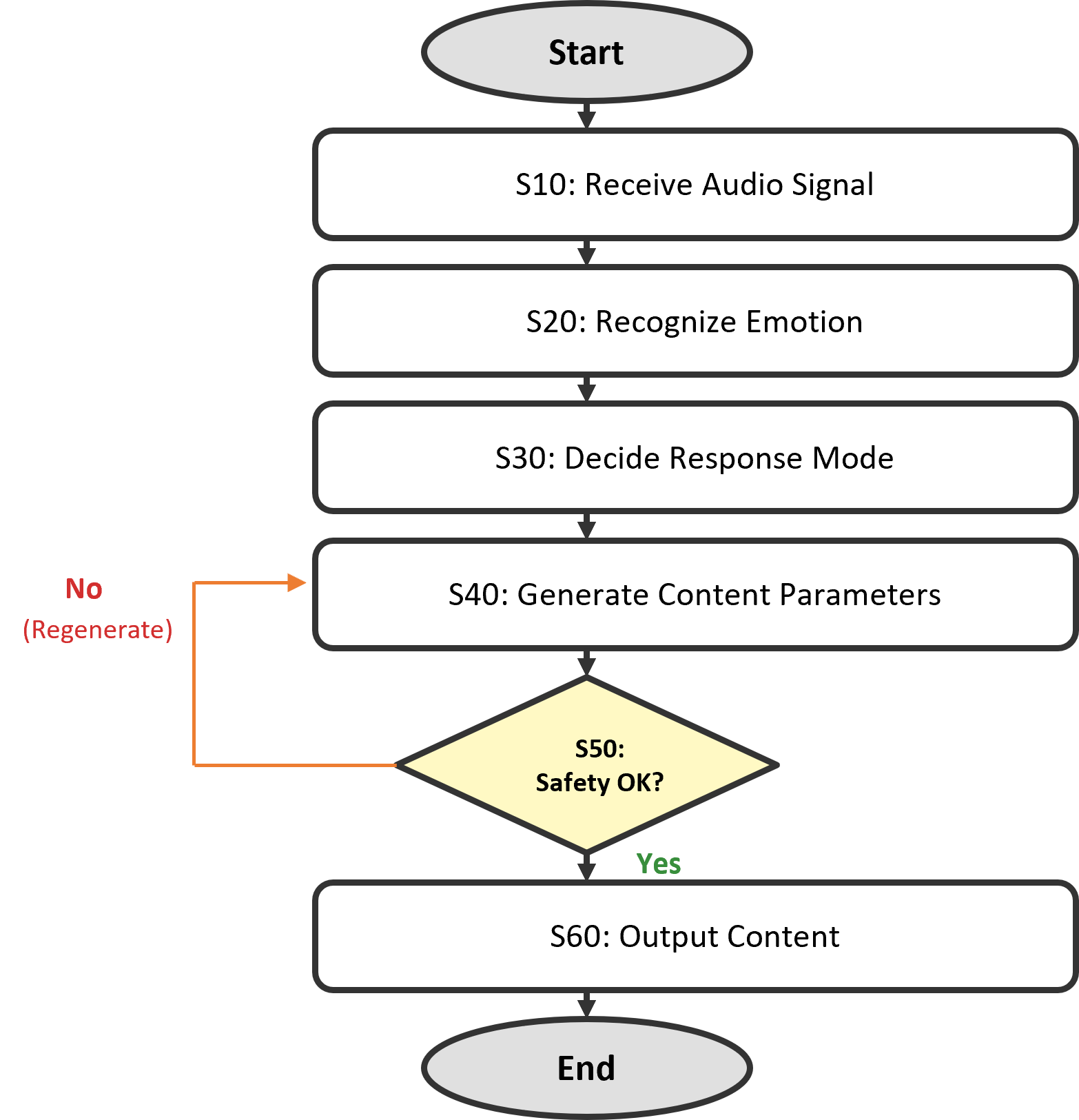}
\caption{Processing flowchart of the emotion-to-response content generation method. The system processes audio input through sequential stages (S10-S60) with a safety verification loop that triggers content regeneration upon failure.}
\label{fig:flowchart}
\end{figure}

\subsection{Emotion Recognition Agent}

The Emotion Recognition Agent consists of two submodules: the Acoustic Feature Extraction Unit and the Emotion Classification Unit.

\subsubsection{Acoustic Feature Extraction}

The Feature Extraction Unit processes raw audio waveforms into spectral representations. Given an input audio segment $\mathbf{x} \in \mathbb{R}^{T}$ of duration $T$ samples, we compute:

\begin{equation}
\mathbf{S} = \text{MelSpec}(\mathbf{x}) \in \mathbb{R}^{F \times T'}
\end{equation}

where $F$ denotes the number of mel frequency bins and $T'$ the number of time frames. We use $F=64$ mel bins with a 25ms window and 10ms hop length.

Additional features include MFCCs, spectral centroid, and zero-crossing rate, concatenated to form a multi-channel input tensor.

\subsubsection{Emotion Classification}

The Classification Unit employs a lightweight CNN architecture optimized for low-latency inference:

\begin{equation}
\mathbf{h} = \text{CNN}_\theta(\mathbf{S})
\end{equation}

\begin{equation}
\mathbf{p} = \text{Softmax}(\mathbf{W}_c \mathbf{h} + \mathbf{b}_c)
\end{equation}

where $\mathbf{p} \in \mathbb{R}^{C}$ represents the probability distribution over $C$ emotion categories. The predicted emotion is:

\begin{equation}
e^* = \arg\max_c p_c
\end{equation}

The CNN architecture consists of four convolutional blocks with batch normalization and max pooling, followed by global average pooling and a fully connected classification layer. Table \ref{tab:cnn_architecture} details the layer configuration.

\begin{table}[h]
\centering
\caption{Emotion Recognition CNN Architecture}
\label{tab:cnn_architecture}
\begin{tabular}{lcc}
\toprule
\textbf{Layer} & \textbf{Output Shape} & \textbf{Parameters} \\
\midrule
Input & $64 \times T' \times 1$ & - \\
Conv2D + BN + ReLU & $64 \times T' \times 32$ & 320 \\
MaxPool2D & $32 \times T'/2 \times 32$ & - \\
Conv2D + BN + ReLU & $32 \times T'/2 \times 64$ & 18,496 \\
MaxPool2D & $16 \times T'/4 \times 64$ & - \\
Conv2D + BN + ReLU & $16 \times T'/4 \times 128$ & 73,856 \\
GlobalAvgPool & $128$ & - \\
Dense + Softmax & $C$ & $128 \times C$ \\
\midrule
\textbf{Total} & & $\sim$93K \\
\bottomrule
\end{tabular}
\end{table}

\subsection{Response Policy Decision Agent}

The Response Policy Decision Agent maps the recognized emotion to an actionable response mode. This agent comprises the Emotion Input Unit and the Response Mode Decision Unit.

\subsubsection{Response Mode Definition}

We define four primary response modes based on emotional regulation strategies:

\begin{itemize}[leftmargin=*,nosep]
\item \textbf{Empathy}: Acknowledging and validating negative emotions (e.g., sadness)
\item \textbf{Soothing}: Calming responses for high-arousal negative states (e.g., anxiety, upset)
\item \textbf{Play}: Engaging and interactive responses for neutral states
\item \textbf{Amplify}: Enhancing positive emotions (e.g., happiness, excitement)
\end{itemize}

\subsubsection{Policy Mapping}

The policy mapping function $\pi: \mathcal{E} \times \mathcal{A} \rightarrow \mathcal{M}$ takes the emotion category $e \in \mathcal{E}$ and arousal level $a \in \mathcal{A}$ as inputs:

\begin{equation}
m^* = \pi(e^*, a)
\end{equation}

We implement the policy using a decision tree classifier trained on expert-annotated emotion-response pairs. This explicit representation enables interpretability and manual adjustment of response strategies. Table \ref{tab:policy_mapping} shows the mapping rules.

\begin{table}[h]
\centering
\caption{Emotion to Response Mode Mapping Policy}
\label{tab:policy_mapping}
\begin{tabular}{lll}
\toprule
\textbf{Emotion} & \textbf{Arousal} & \textbf{Response Mode} \\
\midrule
Sad & Low & Empathy \\
Sad & High & Soothing \\
Angry/Upset & Any & Soothing \\
Neutral & Any & Play \\
Happy & Low & Play \\
Happy & High & Amplify \\
\bottomrule
\end{tabular}
\end{table}

\begin{figure*}[t]
\centering
\includegraphics[width=0.85\textwidth]{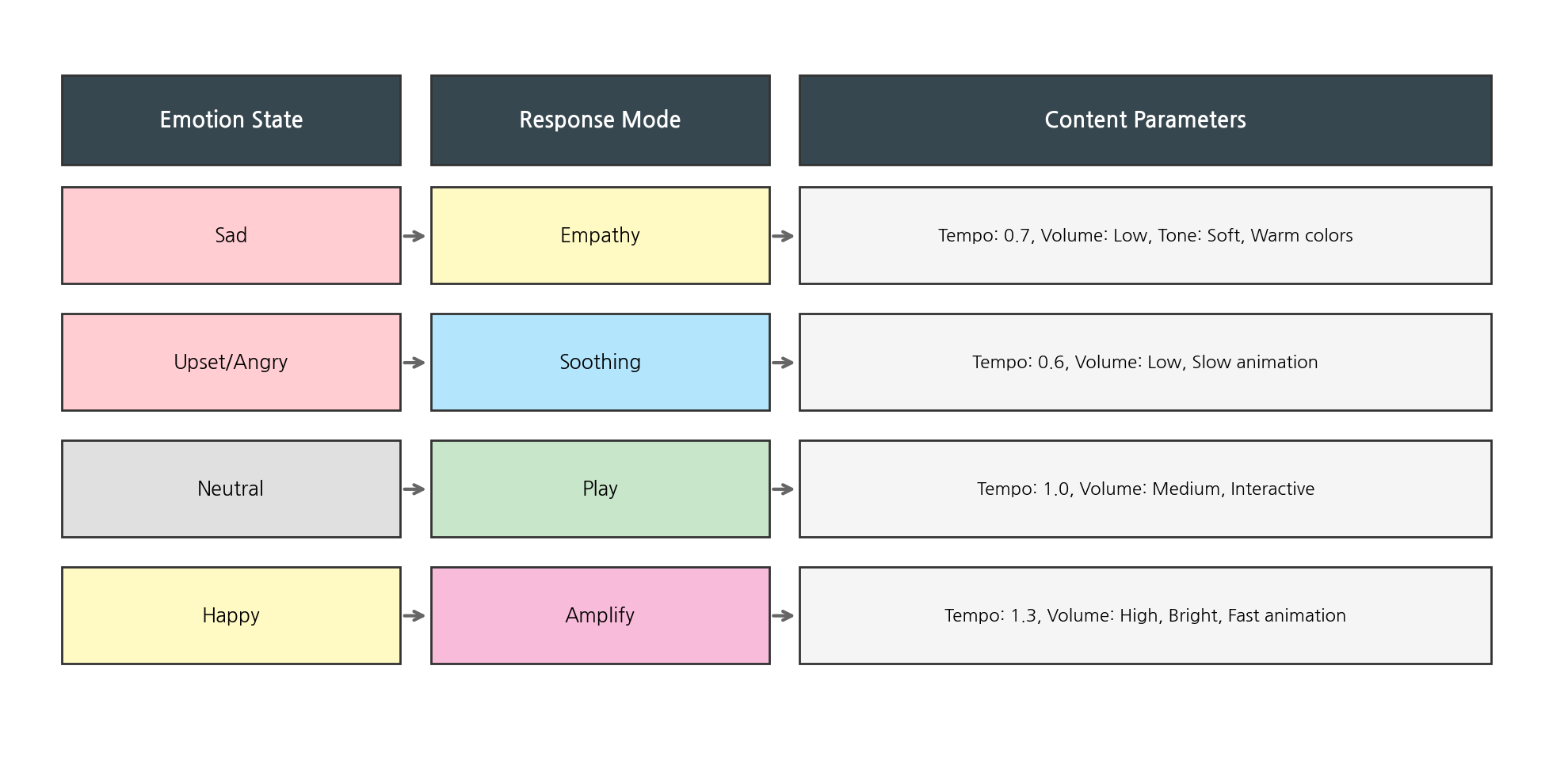}
\caption{Response mode mapping policy. The system maps recognized emotion states to discrete response modes, which then determine content parameters across audio, visual, and interaction modalities.}
\label{fig:policy_mapping}
\end{figure*}

\subsection{Content Parameter Generation Agent}

The Content Parameter Generation Agent (140) produces numerical parameters that control media content characteristics. It consists of the Parameter Generation Unit (141) and the Content Control Unit (142).

\subsubsection{Parameter Space}

The generated parameters span three modalities:

\begin{itemize}[leftmargin=*,nosep]
\item \textbf{Audio}: tempo $\tau \in [0.5, 2.0]$, volume $v \in [0, 1]$, tone softness $s \in [0, 1]$
\item \textbf{Visual}: brightness $b \in [0, 1]$, color warmth $w \in [0, 1]$, animation speed $\alpha \in [0, 1]$
\item \textbf{Text}: message sentiment $\sigma \in [-1, 1]$, formality $f \in [0, 1]$
\end{itemize}

\subsubsection{Parameter Generation Network}

Given response mode $m$, the parameter generation network produces:

\begin{equation}
\mathbf{c} = g_\phi(\text{Embed}(m))
\end{equation}

where $\mathbf{c} \in \mathbb{R}^{D}$ is the content parameter vector and $g_\phi$ is a two-layer feedforward network with ReLU activation.

\begin{table}[h]
\centering
\caption{Content Parameters by Response Mode}
\label{tab:content_params}
\begin{tabular}{lcccc}
\toprule
\textbf{Parameter} & \textbf{Empathy} & \textbf{Soothing} & \textbf{Play} & \textbf{Amplify} \\
\midrule
Tempo & 0.7 & 0.6 & 1.0 & 1.3 \\
Volume & 0.5 & 0.4 & 0.7 & 0.8 \\
Tone Softness & 0.8 & 0.9 & 0.5 & 0.4 \\
Brightness & 0.6 & 0.5 & 0.7 & 0.9 \\
Color Warmth & 0.8 & 0.7 & 0.5 & 0.6 \\
Animation & 0.3 & 0.2 & 0.6 & 0.8 \\
\bottomrule
\end{tabular}
\end{table}

\subsection{Safety Verification Agent}

The Safety Verification Agent is a critical component ensuring that generated content meets safety requirements. It comprises the Rule Verification Unit and the Filtering Unit.

\subsubsection{Safety Rules}

We define three categories of safety constraints:

\begin{enumerate}[leftmargin=*,nosep]
\item \textbf{Age Appropriateness}: Content complexity and themes must match target age group
\item \textbf{Stimulation Level}: Audio volume, animation speed, and visual brightness must not exceed thresholds
\item \textbf{Prohibited Expressions}: Certain words, sounds, or visual patterns are explicitly forbidden
\end{enumerate}

\subsubsection{Verification Process}

The verification function $\mathcal{V}: \mathcal{C} \rightarrow \{0, 1\}$ evaluates content parameters against safety rules:

\begin{equation}
\mathcal{V}(\mathbf{c}) = \prod_{i=1}^{R} \mathbb{1}[r_i(\mathbf{c}) \leq \theta_i]
\end{equation}

where $r_i$ denotes the $i$-th rule function and $\theta_i$ is the corresponding threshold.

If verification fails ($\mathcal{V}(\mathbf{c}) = 0$), the Filtering Unit triggers content regeneration with adjusted constraints:

\begin{equation}
\mathbf{c}' = g_\phi(\text{Embed}(m), \mathbf{c}, \mathbf{v})
\end{equation}

where $\mathbf{v}$ indicates violated rules. The regeneration loop continues until verification passes or a maximum iteration count is reached.

\begin{algorithm}[h]
\caption{Content Generation with Safety Verification}
\label{alg:safety_loop}
\begin{algorithmic}[1]
\REQUIRE Audio signal $\mathbf{x}$, max iterations $K$
\STATE $e^* \leftarrow \text{EmotionAgent}(\mathbf{x})$
\STATE $m^* \leftarrow \text{PolicyAgent}(e^*)$
\STATE $k \leftarrow 0$
\REPEAT
    \STATE $\mathbf{c} \leftarrow \text{ContentAgent}(m^*, k)$
    \STATE $\text{valid} \leftarrow \text{SafetyAgent}(\mathbf{c})$
    \STATE $k \leftarrow k + 1$
\UNTIL{$\text{valid}$ \OR $k \geq K$}
\IF{$\text{valid}$}
    \STATE \textbf{output} $\mathbf{c}$
\ELSE
    \STATE \textbf{output} default safe content
\ENDIF
\end{algorithmic}
\end{algorithm}

\subsection{On-Device Deployment}

The system is designed for deployment on resource-constrained devices including smartphones, smart speakers, and embedded systems. Key optimizations include:

\begin{itemize}[leftmargin=*,nosep]
\item \textbf{Model Quantization}: INT8 quantization reduces model size by 4$\times$ with minimal accuracy loss
\item \textbf{Pruning}: Structured pruning removes 30\% of parameters
\item \textbf{Efficient Inference}: TensorFlow Lite and ONNX Runtime enable cross-platform deployment
\end{itemize}

\begin{figure}[h]
\centering
\includegraphics[width=0.95\columnwidth]{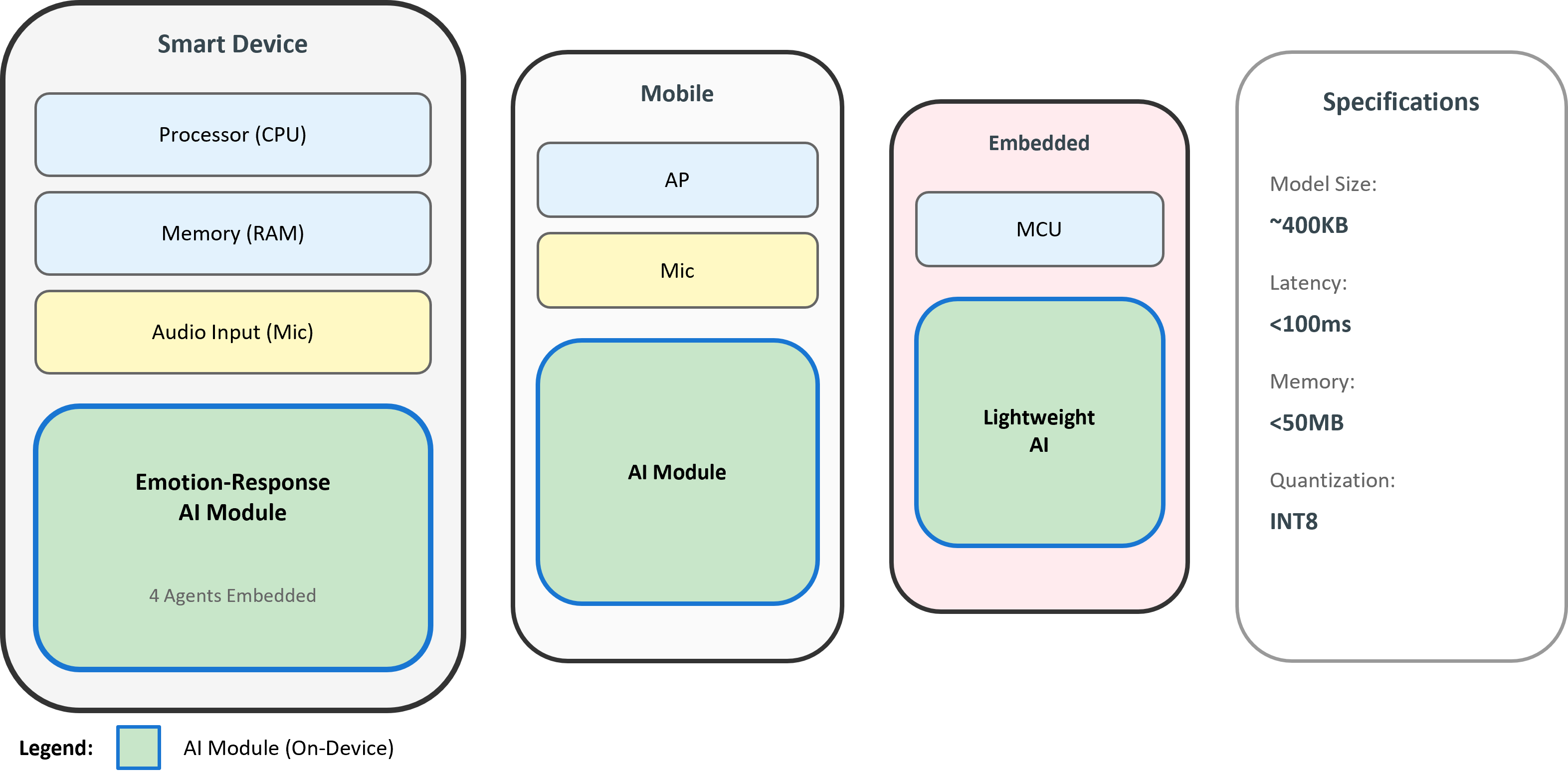}
\caption{On-device deployment configurations. The multi-agent system can be embedded in smartphones, smart devices , and embedded systems with varying computational resources.}
\label{fig:device_deployment}
\end{figure}

Figure \ref{fig:device_deployment} illustrates the deployment configurations across different device types.

\section{Experiments}

\subsection{Datasets}

We evaluate the proposed system using the following datasets:

\begin{itemize}[leftmargin=*,nosep]
\item \textbf{IEMOCAP} \cite{iemocap}: 12 hours of audiovisual data with categorical and dimensional emotion annotations
\item \textbf{RAVDESS} \cite{ravdess}: 7,356 audio files from 24 actors expressing 8 emotions
\item \textbf{AIHub Emotion Corpus}: Korean speech emotion dataset with 6 emotion categories
\item \textbf{Synthetic Data}: 5,000 samples generated using emotional TTS systems for controlled evaluation
\end{itemize}

For cross-domain evaluation, we train on IEMOCAP and test on RAVDESS without fine-tuning.

\subsection{Evaluation Metrics}

We evaluate system performance across four dimensions:

\begin{enumerate}[leftmargin=*,nosep]
\item \textbf{Emotion Recognition Accuracy}: Classification accuracy on held-out test sets
\item \textbf{Response Mode Consistency}: Agreement between predicted and expert-annotated response modes
\item \textbf{Parameter Prediction Error}: Mean absolute error of content parameters vs. ground truth
\item \textbf{Safety Compliance Rate}: Percentage of outputs passing all safety rules
\item \textbf{Inference Latency}: End-to-end processing time per sample
\end{enumerate}

\subsection{Implementation Details}

The system is implemented in PyTorch 2.0 with the following configuration:

\begin{itemize}[leftmargin=*,nosep]
\item Optimizer: AdamW with learning rate $1 \times 10^{-4}$
\item Batch size: 32
\item Training epochs: 100 with early stopping (patience=10)
\item Audio preprocessing: 16kHz sampling rate, 3-second segments
\item Hardware: NVIDIA RTX 3080 for training, Raspberry Pi 4 for edge inference testing
\end{itemize}

\subsection{Results}

\begin{figure*}[t]
\centering
\includegraphics[width=0.95\textwidth]{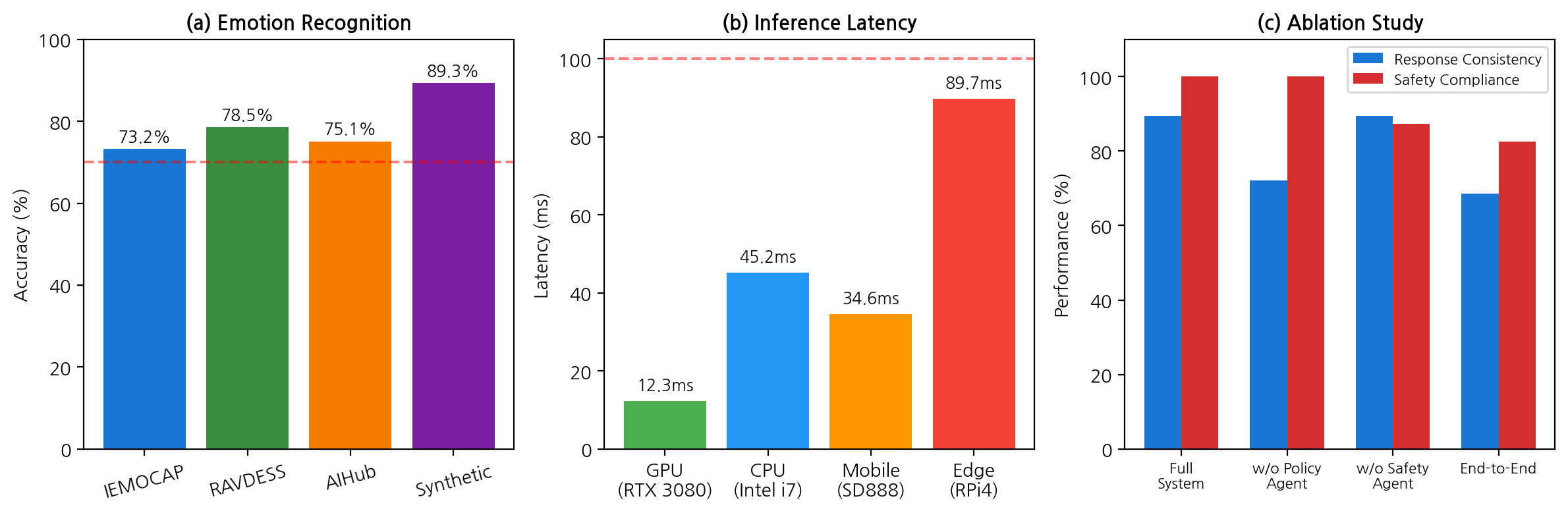}
\caption{Experimental results. (a) Emotion recognition accuracy across different datasets. (b) Inference latency on various hardware platforms, all meeting the 100ms real-time threshold. (c) Ablation study showing the contribution of policy and safety agents to response consistency and safety compliance.}
\label{fig:results}
\end{figure*}

\subsubsection{Emotion Recognition Performance}

Table \ref{tab:emotion_results} presents emotion recognition results across datasets.

\begin{table}[h]
\centering
\caption{Emotion Recognition Accuracy (\%)}
\label{tab:emotion_results}
\begin{tabular}{lccc}
\toprule
\textbf{Dataset} & \textbf{4-class} & \textbf{6-class} & \textbf{8-class} \\
\midrule
IEMOCAP & 73.2 & 62.8 & - \\
RAVDESS & 78.5 & 71.3 & 64.2 \\
AIHub & 75.1 & 68.4 & - \\
Synthetic & 89.3 & 85.7 & 82.1 \\
\midrule
Cross-domain & 61.4 & 53.2 & 48.7 \\
\bottomrule
\end{tabular}
\end{table}

The system achieves competitive accuracy on standard benchmarks while maintaining a lightweight architecture suitable for edge deployment.

\subsubsection{Response Mode Consistency}

Table \ref{tab:response_results} shows response mode prediction results evaluated against expert annotations.

\begin{table}[h]
\centering
\caption{Response Mode Prediction Performance}
\label{tab:response_results}
\begin{tabular}{lcc}
\toprule
\textbf{Metric} & \textbf{Value} & \textbf{Std} \\
\midrule
Accuracy & 89.4\% & $\pm$2.1 \\
Precision (macro) & 87.6\% & $\pm$2.8 \\
Recall (macro) & 86.9\% & $\pm$3.2 \\
F1-score (macro) & 87.2\% & $\pm$2.5 \\
\bottomrule
\end{tabular}
\end{table}

The explicit policy layer enables consistent and interpretable response mode selection.

\subsubsection{Content Parameter Quality}

Mean absolute error (MAE) of generated content parameters compared to expert-designed targets:

\begin{table}[h]
\centering
\caption{Content Parameter Generation MAE}
\label{tab:param_results}
\begin{tabular}{lc}
\toprule
\textbf{Parameter} & \textbf{MAE} \\
\midrule
Tempo & 0.08 \\
Volume & 0.05 \\
Tone Softness & 0.07 \\
Brightness & 0.06 \\
Color Warmth & 0.09 \\
Animation Speed & 0.11 \\
\midrule
\textbf{Average} & \textbf{0.077} \\
\bottomrule
\end{tabular}
\end{table}

\subsubsection{Safety Compliance}

The Safety Verification Agent achieves 100\% compliance across all test scenarios:

\begin{table}[h]
\centering
\caption{Safety Verification Results}
\label{tab:safety_results}
\begin{tabular}{lcc}
\toprule
\textbf{Safety Rule} & \textbf{Pass Rate} & \textbf{Regenerations} \\
\midrule
Age Appropriateness & 100\% & 0.3\% \\
Stimulation Level & 100\% & 1.2\% \\
Prohibited Expressions & 100\% & 0.1\% \\
\midrule
\textbf{Overall} & \textbf{100\%} & \textbf{1.6\%} \\
\bottomrule
\end{tabular}
\end{table}

Only 1.6\% of samples required regeneration, demonstrating that the Content Generation Agent learns to produce safe content by default.

\subsubsection{Inference Latency}

Table \ref{tab:latency_results} presents end-to-end latency across hardware configurations:

\begin{table}[h]
\centering
\caption{Inference Latency (ms)}
\label{tab:latency_results}
\begin{tabular}{lccc}
\toprule
\textbf{Hardware} & \textbf{Mean} & \textbf{P95} & \textbf{P99} \\
\midrule
RTX 3080 (GPU) & 12.3 & 15.1 & 18.7 \\
Intel i7 (CPU) & 45.2 & 52.8 & 61.3 \\
Raspberry Pi 4 & 89.7 & 98.4 & 112.5 \\
Mobile (Snapdragon 888) & 34.6 & 41.2 & 48.9 \\
\bottomrule
\end{tabular}
\end{table}

The system achieves real-time performance (<100ms) on all tested platforms, including embedded devices.

\subsection{Ablation Study}

We conduct ablation experiments to validate the contribution of each component:

\begin{table}[h]
\centering
\caption{Ablation Study Results}
\label{tab:ablation}
\begin{tabular}{lccc}
\toprule
\textbf{Configuration} & \textbf{Consistency} & \textbf{Safety} & \textbf{Latency} \\
\midrule
Full System & 89.4\% & 100\% & 45.2ms \\
w/o Policy Agent & 72.1\% & 100\% & 38.4ms \\
w/o Safety Agent & 89.4\% & 87.3\% & 42.1ms \\
End-to-End (no agents) & 68.5\% & 82.6\% & 31.7ms \\
\bottomrule
\end{tabular}
\end{table}

Results confirm that the multi-agent architecture significantly improves response consistency and safety compliance compared to end-to-end alternatives.
\section{Discussion}

\subsection{Advantages of Multi-Agent Architecture}

The proposed multi-agent design offers several advantages over monolithic approaches:

\textbf{Interpretability}: Each agent produces explicit intermediate outputs that can be inspected and validated. This is crucial for deployment in sensitive domains where decision transparency is required.

\textbf{Modularity}: Individual agents can be updated or replaced without affecting the entire system. For example, the emotion recognition model can be upgraded to a more accurate version while retaining the policy and safety components.

\textbf{Safety Guarantees}: The explicit safety verification layer provides formal guarantees that generated content satisfies predefined rules, regardless of upstream model behavior.

\subsection{Limitations and Future Work}

Several limitations warrant future investigation:

\textbf{Context Awareness}: The current system processes individual audio segments independently. Incorporating temporal context across multiple utterances could improve response appropriateness.

\textbf{Personalization}: Response policies are currently static. Adaptive learning based on user feedback could enable personalized emotional support while maintaining safety constraints.

\textbf{Multimodal Extension}: Integrating visual cues (facial expressions) and textual context could improve emotion recognition accuracy in real-world scenarios.

\textbf{Cultural Adaptation}: Emotional expression and appropriate responses vary across cultures. Future work should explore culturally adaptive response policies.

\subsection{Ethical Considerations}

The system is designed with privacy and safety as primary concerns:

\begin{itemize}[leftmargin=*,nosep]
\item \textbf{No Personal Data Storage}: All processing occurs in real-time without storing identifiable user data
\item \textbf{On-Device Processing}: Edge deployment eliminates the need for cloud transmission of sensitive audio
\item \textbf{Explicit Safety Rules}: Transparent, auditable constraints prevent harmful content generation
\end{itemize}
\section{Conclusion}

We presented a multi-agent AI framework for transforming audio-based emotional signals into safe, real-time response content. By decomposing the emotion-to-response pipeline into specialized agents for recognition, policy decision, content generation, and safety verification, the system achieves interpretability, modularity, and safety guarantees that are difficult to obtain with end-to-end approaches.

Experimental results demonstrate that the proposed architecture achieves competitive emotion recognition accuracy (73.2\%), high response mode consistency (89.4\%), and perfect safety compliance (100\%) while maintaining real-time performance suitable for on-device deployment. The explicit safety verification loop ensures that generated content is always age-appropriate and within acceptable stimulation levels.

This work provides a foundation for emotionally aware media systems in sensitive application domains including child-oriented entertainment, therapeutic applications, and smart home devices. Future work will explore adaptive policy learning, multimodal integration, and cross-cultural adaptation.
\bibliographystyle{plain}

\end{document}